\documentclass[conference]{IEEEtran}

\usepackage{times}
\usepackage{epsfig}
\usepackage{graphicx}
\usepackage{amsmath}
\usepackage{epstopdf}
\usepackage{amssymb}
\graphicspath{{./Images/}}
\usepackage{subfig}
\usepackage{tabularx}
\usepackage{subfig}
\usepackage{array}
\DeclareMathOperator*{\argmin } {arg\, min}

\usepackage{verbatim}
\usepackage{placeins}
\usepackage{algorithm}
\usepackage{algpseudocode}
\newcolumntype{M}[1]{>{\centering\arraybackslash}m{#1}}

\begin{document}

\title{Class Representative Autoencoder for Low Resolution Multi-Spectral Gender Classification}
\author{\IEEEauthorblockN{Maneet Singh, Shruti Nagpal, Richa Singh, and Mayank Vatsa}
\IEEEauthorblockA{IIIT-Delhi, India\\
Email: \{maneets, shrutin, rsingh, mayank\}@iiitd.ac.in}
}

\maketitle

\begin{abstract}
Gender is one of the most common attributes used to describe an individual. It is used in multiple domains such as human computer interaction, marketing, security, and demographic reports. Research has been performed to automate the task of gender recognition in constrained environment using face images, however, limited attention has been given to gender classification in unconstrained scenarios. This work attempts to address the challenging problem of gender classification in multi-spectral low resolution face images. We propose a robust Class Representative Autoencoder model, termed as \textit{AutoGen} for the same. The proposed model aims to minimize the intra-class variations while maximizing the inter-class variations for the learned feature representations. Results on visible as well as near infrared spectrum data for different resolutions  and multiple databases depict the efficacy of the proposed model. Comparative results with existing approaches and two commercial off-the-shelf systems further motivate the use of class representative features for classification.

\end{abstract}

\section{Introduction}

Gender is one of the primary attributes used to describe an individual, which often define and govern the behavioral as well as physical characteristics of a person. Over the past several decades, researchers have attempted to understand the behavioral differences between the two genders \cite{gender1, gender2}. These behavioral difference have further been explored for enhancing digital user experience, human computer interaction, and gender targeted advertisements. On the other hand, human beings have always used the physical differences as a key identifying attribute of an individual. This has resulted in utilization of gender information in scenarios of surveillance, security, and access control. Given the tremendous applications of automated gender classification, researchers have proposed several novel algorithms using different modalities to model the same \cite{faceGender, irisGender, fingerprintGender, keystrokeGender, gaitGender}. 

Face of an individual is one of the most distinguishable and non-invasive modalities for gender recognition. Automated gender recognition\footnote{Gender classification and gender recognition have been used interchangeably.} on face images has attracted the attention of researchers since a long time. While gender recognition in controlled, well-illuminated scenarios has been well explored, it is still considered an arduous task in unconstrained scenarios. For example, low resolution faces captured in uncontrolled surveillance settings are difficult to process by most of the existing automatic gender classification algorithms.      

\begin{figure}
\includegraphics[width=\linewidth]{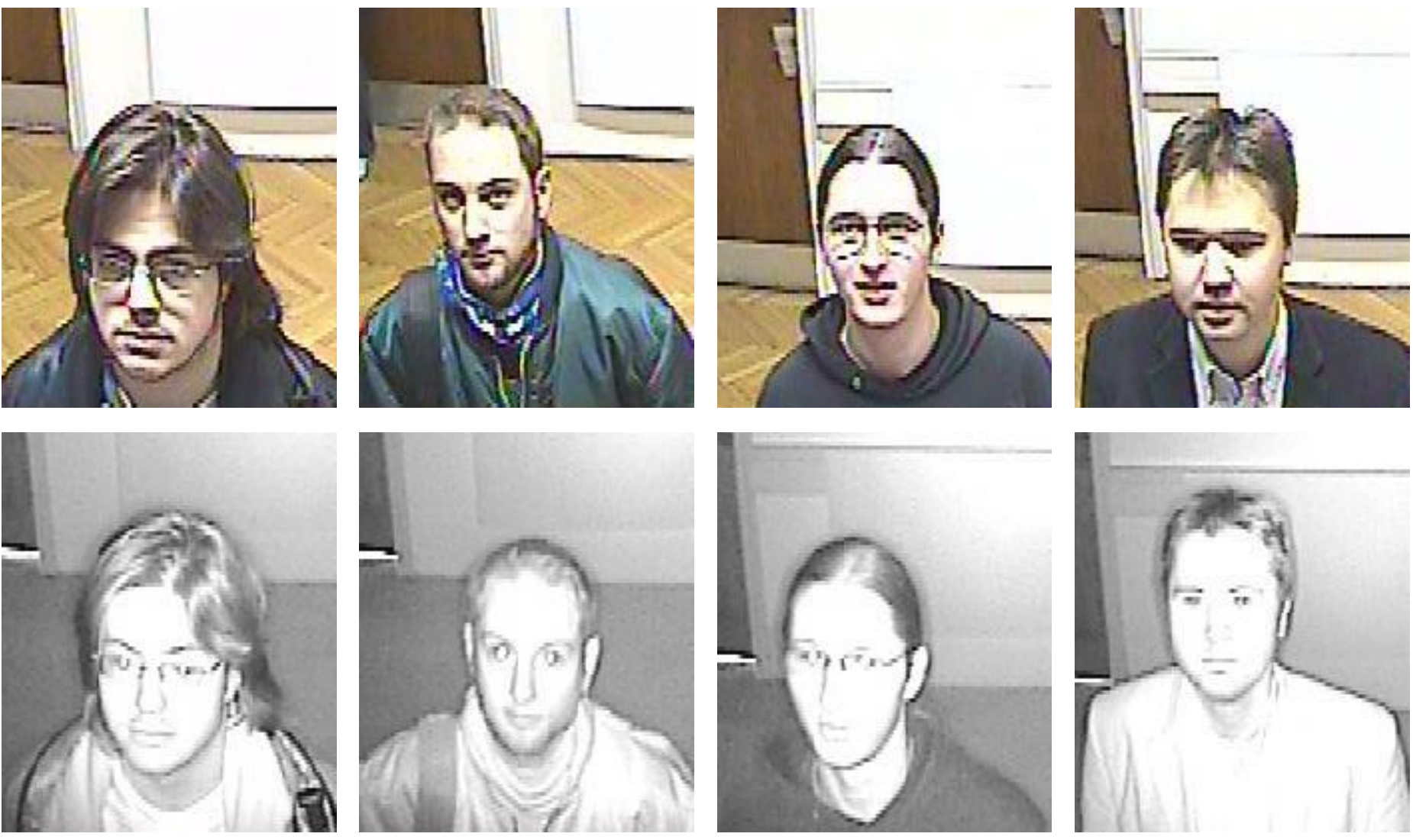}
\caption{Sample male images from SCface dataset \cite{scface}. Each column corresponds to one subject. The top row displays images captured in the visible domain, while the second row depicts images captured in the NIR spectrum. }
\label{intro}
\end{figure}

In general, images captured from surveillance (or CCTV) cameras entail non-cooperative subjects in unconstrained environments. These images are often of poor quality, resulting in low resolution face regions. Moreover, the videos/images are captured in different illumination conditions, during day and night time. Most surveillance cameras have dual capabilities of capturing near infrared (NIR) spectrum images in the night time, along with RGB/grayscale images during the day time. Some sample images captured in both the spectra (visible and NIR) from surveillance cameras are presented in Fig. \ref{intro}. It can be observed that low resolution face images often contain less information content along with several challenging covariates. These challenges require developing a robust algorithm which is capable of performing gender classification for low resolution images, in both visible and NIR spectrum.

In literature, researchers have explored several techniques to address the task of gender recognition in visible spectrum. Moghaddam \textit{et al.} \cite{moghaddam02} presented the superior performance of non-linear Support Vector Machines for performing gender recognition in low-resolution thumbnail images. Comparative analysis with other techniques such as Fisher linear discriminant and nearest neighbor classifiers presented the strength of their proposed approach. Andreu \textit{et al.} \cite{andreu14} varied the resolution of face images from $2\times1$ to $329\times264$ pixels and studied it's effect on gender recognition for large datasets. Experiments were performed on pixel intensity values with two classifiers, and reduced performance was observed with low resolution faces images. In 2016, Juefei-Xu \textit{et al.} \cite{deepGender} proposed DeepGender, a progressive convolutional neural network training technique for the task of gender recognition, with an application to low resolution face images. The model aimed to learn important regions of the face for the given task and yielded promising results. 

Recently, researchers have also focused upon understanding and improving the performance of gender classification in the NIR spectrum as well. Chen and Ross \cite{ross11} proposed to perform gender classification on NIR spectrum face images of size $130\times150$ using Local Binary Patterns (LBP) and multiple classifiers. Subsequently, Ross and Chen \cite{ross11_2} evaluated the performance of gender classification across resolutions in cross spectrum scenarios (visible to NIR and vice-versa) by using features extracted via Principal Component Analysis, and Support Vector Machines for classification. Recently, Narang and Bourlai \cite{narang16} proposed a deep convolutional neural network based architecture for NIR spectrum face images (dimension $128 \times 128$). The existing literature showcases that NIR images contain discriminative information for performing gender classification. However, the degree of class-variation is noticeably less, as compared to the visible spectrum.

This research aims to address the task of gender classification in low resolution images, for both visible as well as NIR spectrum. The contributions of this research are (i) proposing a novel formulation of Class Representative Autoencoder which encodes class-specific features, and (2) utilizing \textit{Autogen}, a Class Representative Autoencoder model for gender classification on low resolution face images (in both visible and NIR spectrum). During the feature learning process, AutoGen aims to incorporate the inherent characteristics of male and female facial features. Using low resolution face images in both visible and NIR spectrum, experimental results and comparison with existing approaches illustrate the efficacy of the proposed approach. The remainder of this paper is organized as follows: the next section describes the proposed formulation and the gender classification algorithm. Section \ref{db} gives the experimental protocols and datasets used, followed by the results and conclusions of this research. 
 
\section{Proposed Algorithm}
Inspired by the characteristics of representation learning, wherein the learned model encodes variations spanning across the training set resulting in greater generalization properties, in this research, a Class Representative Autoencoder is proposed. Gender classification is performed using the proposed model, termed as \textit{AutoGen}. The model performs gender recognition across multiple spectrum on low resolution face images. It is built upon a traditional autoencoder and aims to model the \textit{inter-class} and \textit{intra-class} variations during feature extraction. 

\subsection{Unsupervised Autoencoder}
Auoencoders are neural networks traditionally used to learn representations for the given data in an unsupervised manner. For a given sample ${x}$, hidden feature representation (${r_x}$) is learned such that the difference between the input data ${x}$  and reconstructed data ${\hat{x}}$ is minimized. 
The objective of the model is to learn weight parameters in order to produce a representation which preserves the non-linearities present in the data, and disregard the redundant information. The loss function for a single layer autoencoder is given by:
\begin{equation} \label{basic}
{
\min \left \|x -  \hat{x} \right \|_{F}^{2} 
}
\end{equation}
The reconstructed sample, ${\hat{x}}$ can be expressed as a function of the input sample, ${\hat{x}} = g \circ f({x})$. Updating Eq. \ref{basic}, the loss function can be rewritten as:
\begin{equation} \label{eqLossAE}
{
\argmin_{\textit{$\mathbf{W}_e,\mathbf{W}_d$}}  \left \|x -  g \circ f(x) \right \|_{F}^{2} 
}
\end{equation}
where, $f({x})$ corresponds to the hidden representation, such that $f({x})$ = $\phi(\mathbf{W}_ex)$. $\mathbf{W}_e$ corresponds to the encoding weights, and $\phi$ is any linear or non-linear function, such as $sigmoid$ or $tanh$. The reconstruction of the sample, ${\hat{x}}$ is computed as $g\circ f({x})$ = $\mathbf{W}_d(f({x}))$, where $\mathbf{W}_d$ corresponds to the decoding weights. The reconstructed data can thus be computed as $\mathbf{W}_d (\phi(\mathbf{W}_{e}x))$. The hidden representation is the learned feature vector which is used for classification. In an attempt to learn finer features, researchers have proposed taking this model \textit{deeper} by introducing the Stacked Autoencoder (SAE) \cite{sdae}. SAE consists of stacked encoder-decoder layers with greedy layer by layer training which enables learning of higher level representative features capturing the higher level of non-linearities. Further extensions are proposed including sparse autoencoders using KL-divergence and $\ell_1$-norm \cite{ng2011sparse}. 


\subsection{Proposed Class Representative Autoencoder}
Recently, \textit{supervised} extensions of traditional unsupervised model of the autoencoder have also been proposed \cite{highContract, contrast, super, pamiMaj}. Most of these algorithms incorporate class information at the time of feature extraction with an aim to reduce only the intra-class variations. On the other hand, the proposed Class Representative Autoencoder is built over the traditional unsupervised autoencoder, modeling both inter-class and intra-class variations at the time of feature learning. The optimization function incorporates class-specific terms for discriminative feature learning, in order to learn class-specific representations. This is done by minimizing the distance between a sample's representation and the representative feature of the sample's class, while maximizing the distance from the representative feature of all other classes. The class representative feature is computed as the mean feature vector of all the samples of a given class. The additional terms aid in incorporating inter-class and intra-class variations during the feature learning process such that the learned features are discriminative in nature. In a $n$ class problem, for a sample ${x_s}$ belonging to class $s$, the proposed model can be written as:
\begin{multline} \label{AutoGen}
\begin{gathered}
\argmin_{\textit{$\theta$}}  \left \| {x_s} -  g \circ f({x_s}) \right \|_{F}^{2} + \lambda_s\left \|{r_{x_{s}}} - {mean_{s}} \right \|_{F}^{2} \\
- \sum_{i = 1}^{n} \lambda_i\left \|{r_{x_{s}}} - {mean_{i}} \right \|_{F}^{2}; \ \ \ \forall i \neq s
\end{gathered}
\end{multline}
where, $\theta = \{\mathbf{W}_d, \mathbf{W}_e\}$, ${r_{x_{s}}}$ refers to the hidden representation of sample ${x_s}$, and $\lambda_s$ and $\lambda_i$ refer to the regularization constants for the additional terms. ${mean_i}$ refers to the mean feature (hidden) representation of all samples belonging to class $i$. For a single layer AutoGen, it can be computed as:
\begin{equation} \label{mean}
{mean_{i}} =  \mu(\phi(\mathbf{W}_{e}\mathbf{X}_{i}))
\end{equation}
where, $\mathbf{W}_e$ are the encoding weights, $\mathbf{X}_i$ corresponds to all the training samples belonging to the $i^{th}$ class, and $\mu$ refers to the mean operator. Expanding Eq. \ref{AutoGen}:
\begin{equation} \label{AutoGenElaborate}
\begin{gathered}
\argmin_{\textit{$\theta$}}  \left \| {x_s} -  \mathbf{W}_d\phi(\mathbf{W}_ex_s) \right \|_{F}^{2}  \\
+ \lambda_s\left \|\phi(\mathbf{W}_ex_s) - \mu(\phi(\mathbf{W}_e\mathbf{X}_s)) \right \|_{F}^{2} \\
- \sum_{i = 1}^{n} \lambda_i\left \|\phi(\mathbf{W}_ex_s) - \mu(\phi(\mathbf{W}_e\mathbf{X}_i)) \right \|_{F}^{2} \forall i \neq s
\end{gathered}
\end{equation}

Thus, the proposed formulation (Eq. \ref{AutoGenElaborate}) consists of three terms: a term for learning the sample's feature, second for incorporating class similarity, and third for incorporating dissimilarity with other classes. As is the case with the traditional autoencoder, the first term aims to reduce the reconstruction error. The second term is responsible for bringing the learned representation (${r_{x_{s}}}$) of a sample ${x}$ belonging to the class $s$  closer to ${mean_s}$ (representative feature of class $s$). Since the entire loss function is minimized, this \textit{intra-class} term is also minimized, thus resulting in learning feature vectors closer to the mean feature representation of that class. The third term in Eq. \ref{AutoGen} is responsible for maximizing the distance between the learned feature vector from the representative features of all other classes. This \textit{inter-class} term attempts to force the model to learn a feature representation different from the representative (mean) feature of all other classes, thereby incorporating discriminability in the feature learning process.
\begin{figure}
\centering
\subfloat[Male]{\includegraphics[trim={2cm 1.5cm 2cm 0.7cm},clip, width= 1in]{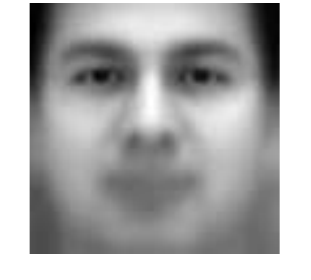}} 
\subfloat[Female]{\includegraphics[trim={2cm 1.5cm 2cm 0.7cm},clip, width= 1in]{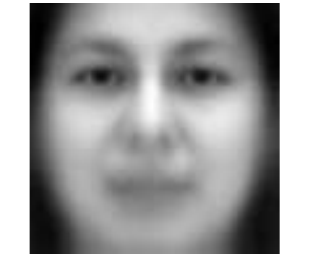}}  
\caption{Mean-male and mean-female images obtained from CMU Multi-PIE dataset \cite{mpie}.}
\label{mean}
\end{figure}

\begin{figure}[t]
\centering
\includegraphics[trim={0.1cm 0.1cm 0cm 0.2cm},clip, width= 3.6in]{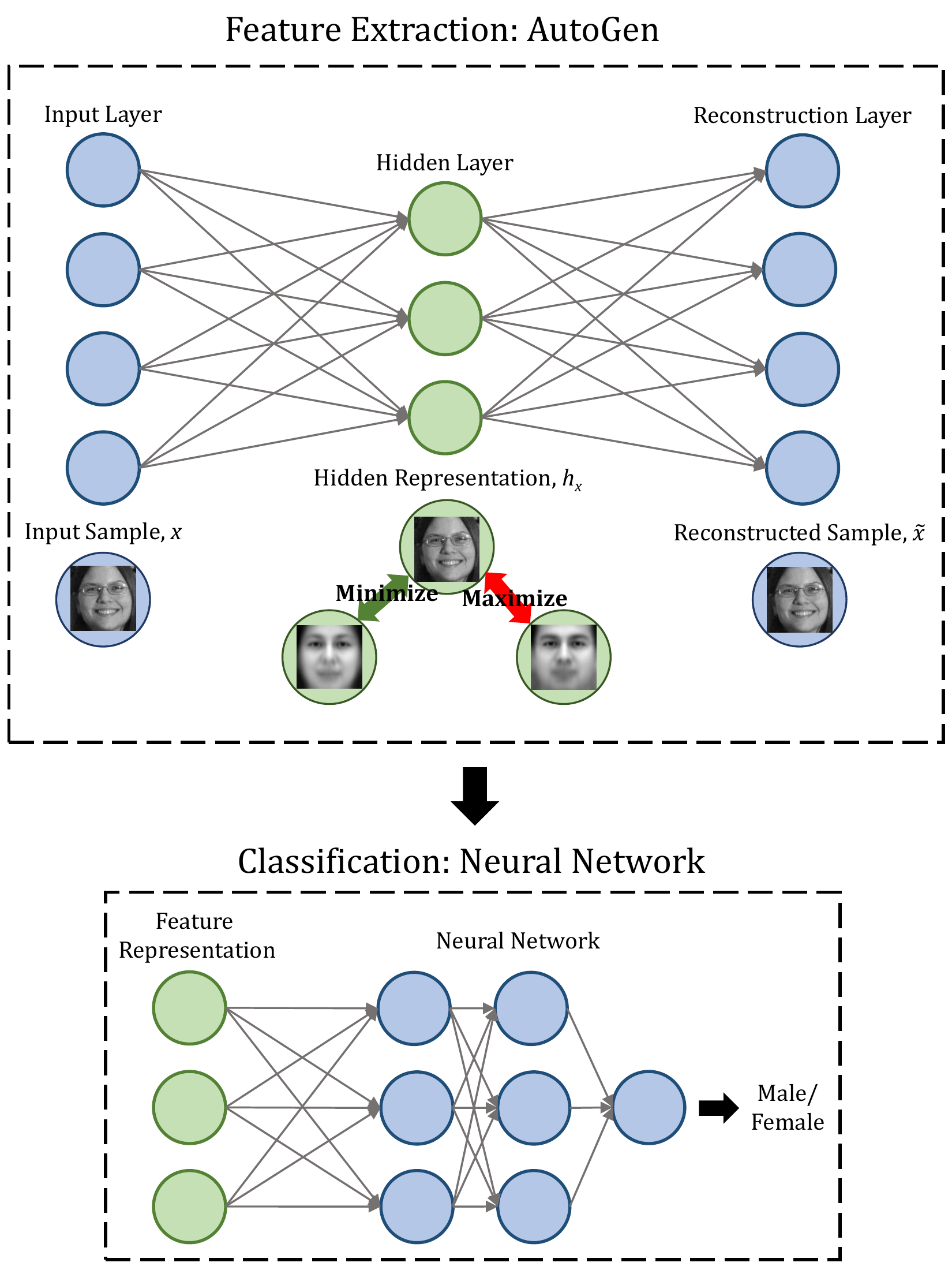}
\caption{Pipeline for performing gender recognition on low resolution face images. The proposed AutoGen model is used for feature extraction, followed by a neural network for classification. AutoGen aims to learn discriminative features by incorporating inter-class and intra-class variations during feature learning. }
\label{algo}
\end{figure}

In order to train the above model, back-propagation is performed using the gradient descent approach \cite{backProp}. The model is trained for $k$ iterations, such that at the ${j^{th}}$ iteration, the parameters learned in the previous $({j-1)^{th}}$ iteration are updated using the derivative of the loss function. The weight update rule can be written as:
\begin{equation} \label{weightUpdate}
\mathbf
{
W^{j} = W	^{j-1} - \eta \frac{\partial E}{\partial W}
}
\end{equation}
where, $\mathbf{E}$ corresponds to the loss function given in Eq. \ref{AutoGenElaborate}. Since all three terms in the formulation are differentiable (when using a differentiable activation function), the rule for updating the weights at each iteration can be obtained by calculating the derivative of each term. In order to extend the algorithm for deeper layers, Class Representative Autoencoder can be trained in a greedy layer-by-layer manner \cite{greedy}. 

\subsection{AutoGen: Class Representative Autoencoder for Visible and NIR Low Resolution Gender Recognition} 

As shown in Fig. \ref{mean}, a significant difference can be observed visually between the \textit{mean-male} and \textit{mean-female} images. This illustrates the need for learning class-specific features for improved classification performance. The proposed formulation aims to learn discriminative features in order to enhance classification of a given face sample into one of the two classes - male or female. At the time of training, mean representations for both the classes are calculated and incorporated by AutoGen for feature learning. The loss function, for a male input sample can be written as follows:
\begin{multline} \label{male}
\begin{gathered}
\argmin_{\textit{$\mathbf{\theta}$}}  \left \| {x_{m}} -  g \circ f({x_{m}}) \right \|_{F}^{2} + \lambda_m\left \|{r_{x_{m}}} - {mean_{m}} \right \|_{F}^{2} \\
- \lambda_f\left \|{r_{x_{m}}} - {mean_{f}} \right \|_{F}^{2}
\end{gathered}
\end{multline}
where, $r_{x_{m}}$ is the learned representation for the male input sample $x_{m}$, and $mean_{m}$ is the mean of features of all the male samples. $mean_{f}$ corresponds to the mean of feature vector of all the female samples. Similarly, the loss function for a female input sample is as follows:
\begin{multline} \label{female}
\begin{gathered}
\argmin_{\textit{$\mathbf{\theta}$}}  \left \| {x_{f}} -  g \circ f({x_{f}}) \right \|_{F}^{2} + \lambda_f\left \|{r_{x_{f}}} - {mean_{f}} \right \|_{F}^{2} \\
- \lambda_m\left \|{r_{x_{f}}} - {mean_{m}} \right \|_{F}^{2}
\end{gathered}
\end{multline}
where, $r_{x_{f}}$ is the learned representation for the female input sample $x_{f}$. The additional terms in the proposed model contribute to the supervised regularization during the feature learning process. The proposed model is used for feature extraction, which is followed by a neural network for classification. Fig. \ref{algo} illustrates a pictorial representation of the proposed algorithm. At the time of testing, the learned encoding weights of AutoGen ($\mathbf{W}_e$) are used to obtain the feature representation of the given test face image. The feature is then provided as input to the trained neural network, which finally predicts whether the input face image belongs to a \textit{male} class or a \textit{female} class. 

\section{Datasets and Experimental Protocol}
\label{db}
This research aims to address gender classification in low resolution face images when images are captured in visible and NIR spectrum. In order to evaluate the performance of the proposed algorithm, three datasets are used that contain images at different resolutions. 
\subsection{Datasets Used}
\begin{figure}
\centering
\includegraphics[width=1\linewidth]{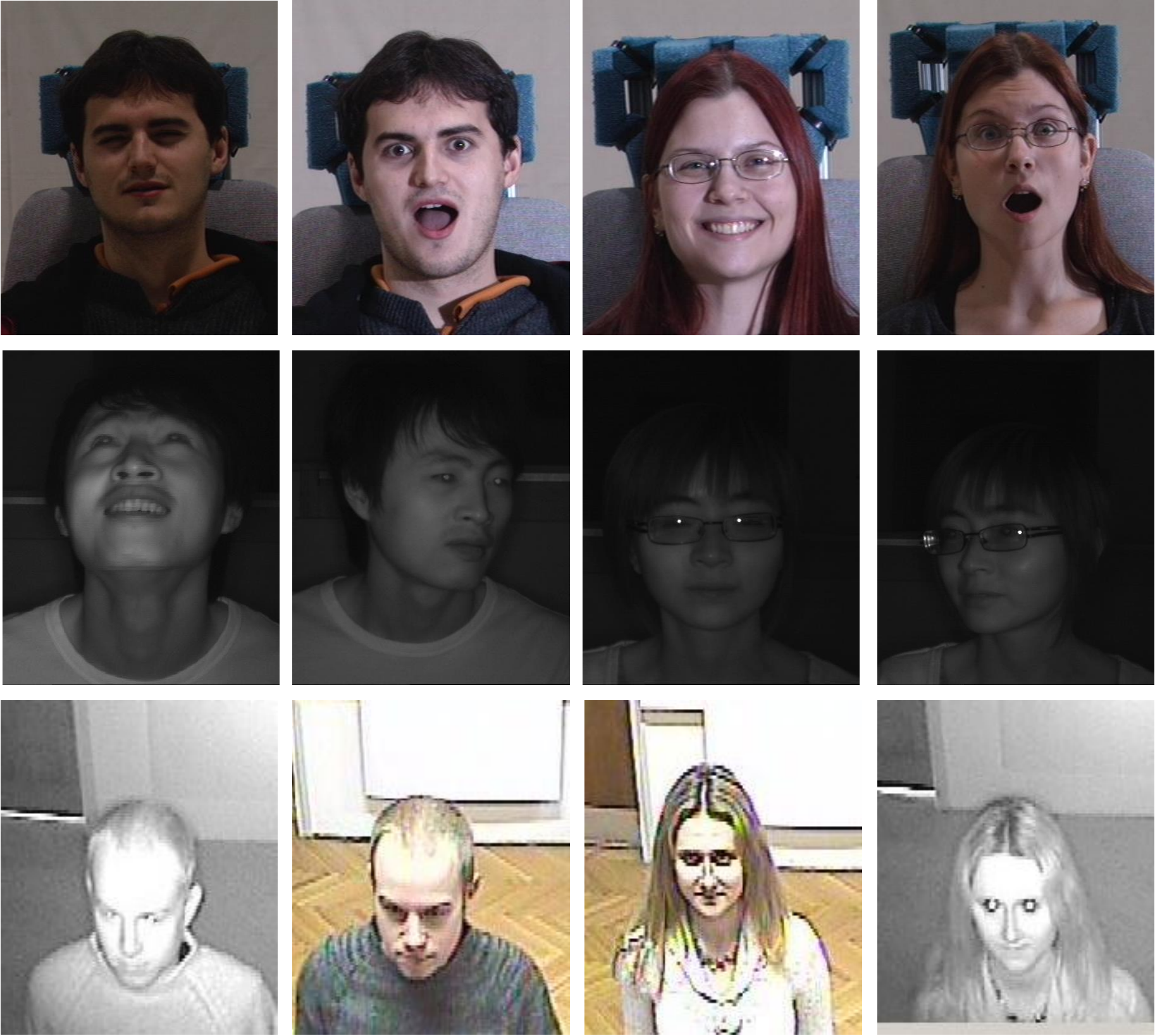}
\caption{Sample images from the datasets used for experimental evaluation. First row corresponds to CMU Multi-PIE dataset \cite{mpie}, second to PolyU-NIR \cite{polyu}, and the last row refers to the SCface dataset \cite{scface}. Each row depicts images pertaining to two subjects.}
\label{sampleDb}
\end{figure}
Experiments are performed on CMU Multi-PIE \cite{mpie}, PolyU-NIR \cite{polyu}, and SCface \cite{scface} databases, and Fig. \ref{sampleDb} displays sample images from each dataset. Details of these databases are as follows: 

\textbf{CMU Multi-PIE Dataset \cite{mpie}:} The dataset contains images of 337 subjects, captured in indoor settings having variations with respect to pose, illumination, and expression. A frontal only subset of the dataset containing 50,248 images is used in the experiments. The dataset is divided into training and testing partitions, such that the train set consists of 18,420 images and the remaining 31,828 images make up the test set. Equal samples from both classes is ensured in the training partition. 

\textbf{PolyU-NIR \cite{polyu}:} The PolyU NIR Face Dataset consists of 34,000 NIR spectrum face images of 335 subjects, having pose, illumination and distance variations. Out of these, 9,960 images are used for training (4,980 per class) and the remaining images form the test set. 

\textbf{SCface Dataset \cite{scface}:} This dataset consists of visible and NIR images of 130 subjects captured over three distances. For each distance, one subject has five visible images and two NIR images captured in uncontrolled indoor environment. Since the aim is to perform low-resolution gender recognition, images pertaining to only the farthest distance have been used in the experiments. For visible spectrum experiments, 100 images are used for training, and 550 are used for testing. For NIR spectrum experiments, 52 images are used for training and the remaining form the test set. Exclusivity of subjects across the training and testing partition is maintained. 

\begin{table}
\renewcommand*{\arraystretch}{1.2}
\begin{center}
\centering
\caption{Classification accuracies (\%) for gender classification on $24\times24$ face images on two datasets: CMU Multi-PIE and PolyU-NIR face datasets. }
\begin{tabular}{|M{8em}||M{7.5em}|M{7.5em}|} 
\hline
\textbf{Algorithm} & \textbf{CMU Multi-PIE} & \textbf{PolyU-NIR} \\
\hline
\hline
 AE & 88.80 & 63.46  \\ 
 \hline
DAE &  88.21 & 62.95  \\
\hline
DBN & 79.99 & 65.10 \\
\hline
DRBM & 72.84 & 50.01  \\
\hline
Face++ &  73.93 & 52.46  \\
\hline
Luxand & 74.14 & 35.10 \\
\hline
\textbf{AutoGen} & \textbf{90.10} & \textbf{71.32} \\
\hline
\end{tabular}
\label{res24}
\end{center}
\end{table}

\subsection{Experimental Protocol}
As part of pre-processing, face detection is performed on all datasets using Viola Jones face detector \cite{viola}, which is followed by geometric normalization of the face images. The detected faces are then resized to specific resolutions for different experiments. For all the datasets, the training partition is used to train the feature learning models and classifier, while testing is performed on the test set. The following experiments are performed for performance evaluation: \\\textbf{(A) Gender Recognition on Face Images with $24 \times 24$ Resolution:} 
\begin{itemize}
\item Visible Spectrum
\begin{itemize}
\item CMU Multi-PIE Dataset: Experiments are performed using the training (18,420 images) and testing partitions (31,828 images) described earlier.
\item SCface Dataset: Only the visible images (training: 100, testing: 550) are used in this experiment. Since the number of samples are less, feature extraction models trained on CMU Multi-PIE dataset are \textit{fine-tuned} using the training set of SCface dataset.
\end{itemize}
\item NIR Spectrum
\begin{itemize}
\item PolyU-NIR Dataset: Evaluation is performed using the training (9,960 images) and testing partitions specified earlier.   
\item SCface Dataset: As was the case with the visible domain, due to lack of training samples, \textit{fine-tuning} is performed on the learned model of PolyU-NIR using the training partition of SCface dataset.
\end{itemize}
\end{itemize}
\textbf{(B) Gender Recognition on Face Images with $16 \times 16$ Resolution:} 
\begin{itemize}
\item Visible Spectrum
\begin{itemize}
\item CMU Multi-PIE Dataset: The same protocol is followed as was done with $24\times24$ resolution images. 
\end{itemize}
\item NIR Spectrum
\begin{itemize}
\item PolyU-NIR Dataset: The same training-testing division was utilized as was for experiments with $24\times24$ resolution face images earlier.
\end{itemize}
\item Spectrum-Invariant Gender Classification
\begin{itemize}
\item The performance of a single AutoGen model for both NIR, as well as visible images is analyzed. In this case the training sets of CMU Multi-PIE and PolyU-NIR datasets are combined to create a single multi-spectrum training set. Gender classification performance is reported with the proposed AutoGen model on the testing partitions of the two datasets.   
\end{itemize}
\end{itemize}

\section{Experiments and Results}

In order to compare the performance of the proposed model with existing algorithms, comparison has been drawn with Autoencoders (AE) \cite{ae}, Denoising Autoencoders (DAE) \cite{dae}, Deep Belief Networks (DBN) \cite{dbn}, Discriminative Restricted Boltzmann Machine (DRBM) \cite{drbm}, and two Commercial-Off-The-Shelf (COTS) systems, Luxand \cite{luxand} and CNN-based Face++ \cite{faceplusplus}. All models are trained with a fixed architecture of $[k, k]$, where $k$ is the size of the input image. Once the feature learning process is over, features are extracted using the model and a Neural Network is trained for the 2-class classification problem of male versus female. A two layer neural network of dimensionality $[l, \frac{l}{4}, \frac{l}{16}]$ is trained, where $l$ is the length of the feature vector. $sigmoid$ activation function was used on all hidden layers. Due to the large class-imbalance of test samples, mean class-wise accuracy is reported for all the experiments. 

\subsection{Classification Performance on CMU Multi-PIE and PolyU-NIR Datasets}
$\mathbf{24\times24:}$ Table \ref{res24} gives the classification accuracies obtained for $24\times24$ face images for visible and NIR spectrum. For both the spectra, AutoGen outperforms existing algorithms and commercial-off-the-shelf systems by reporting higher classification accuracies. In case of CMU Multi-PIE, AutoGen gives a classification accuracy of 90.10\%, which is at least 7\% higher than COTS (Face++). For PolyU-NIR face dataset, AutoGen presents a classification accuracy of 71.32\%, which gives an improvement of at least 7\% over existing algorithms, and at least 20\% over COTS. 

\begin{table}
\renewcommand*{\arraystretch}{1.2}
\begin{center}
\centering
\caption{Classification accuracies (\%) for gender classification on $16\times16$ face images from CMU Multi-PIE and PolyU-NIR face datasets. }
\begin{tabular}{|M{8em}||M{7.5em}|M{7.5em}|} 
\hline
\textbf{Algorithm} & \textbf{CMU Multi-PIE} & \textbf{PolyU-NIR} \\
\hline
\hline
 AE & 88.80 & 61.28  \\ 
 \hline
DAE &  87.82 & 63.91  \\
\hline
DBN &  72.36 & 49.00 \\
\hline
DRBM & 70.41 & 52.93  \\
\hline
Face++ &  0.00 & 0.00  \\
\hline
Luxand & 0.00 & 0.00 \\
\hline
\textbf{AutoGen} & \textbf{89.57} & \textbf{69.86} \\
\hline
\end{tabular}
\label{res16}
\end{center}
\end{table}

\begin{table}
\renewcommand*{\arraystretch}{1.2}
\begin{center}
\centering
\caption{Class specific classification accuracies (\%) obtained with \textbf{AutoGen} for gender classification.}
\begin{tabular}{|M{8em}||M{7.5em}|M{7.5em}|} 
\hline
\textbf{Dataset} & \textbf{Male} & \textbf{Female} \\
\hline
\hline
\multicolumn{3}{|c|}{$24\times24$ Face Images}     \\
\hline
\hline
CMU Multi-PIE & 87.47 & 92.73  \\ 
 \hline
PolyU-NIR &  66.47 & 76.17  \\
\hline
\multicolumn{3}{|c|}{$16\times16$ Face Images}     \\
\hline
\hline
CMU Multi-PIE & 86.50 & 92.64  \\ 
 \hline
PolyU-NIR &  70.47 & 69.24  \\
\hline
\end{tabular}
\label{classWise}
\end{center}
\end{table}

\begin{figure*}[h]
\centering
\subfloat[CMU Multi-PIE]{\includegraphics[width= 3.6in]{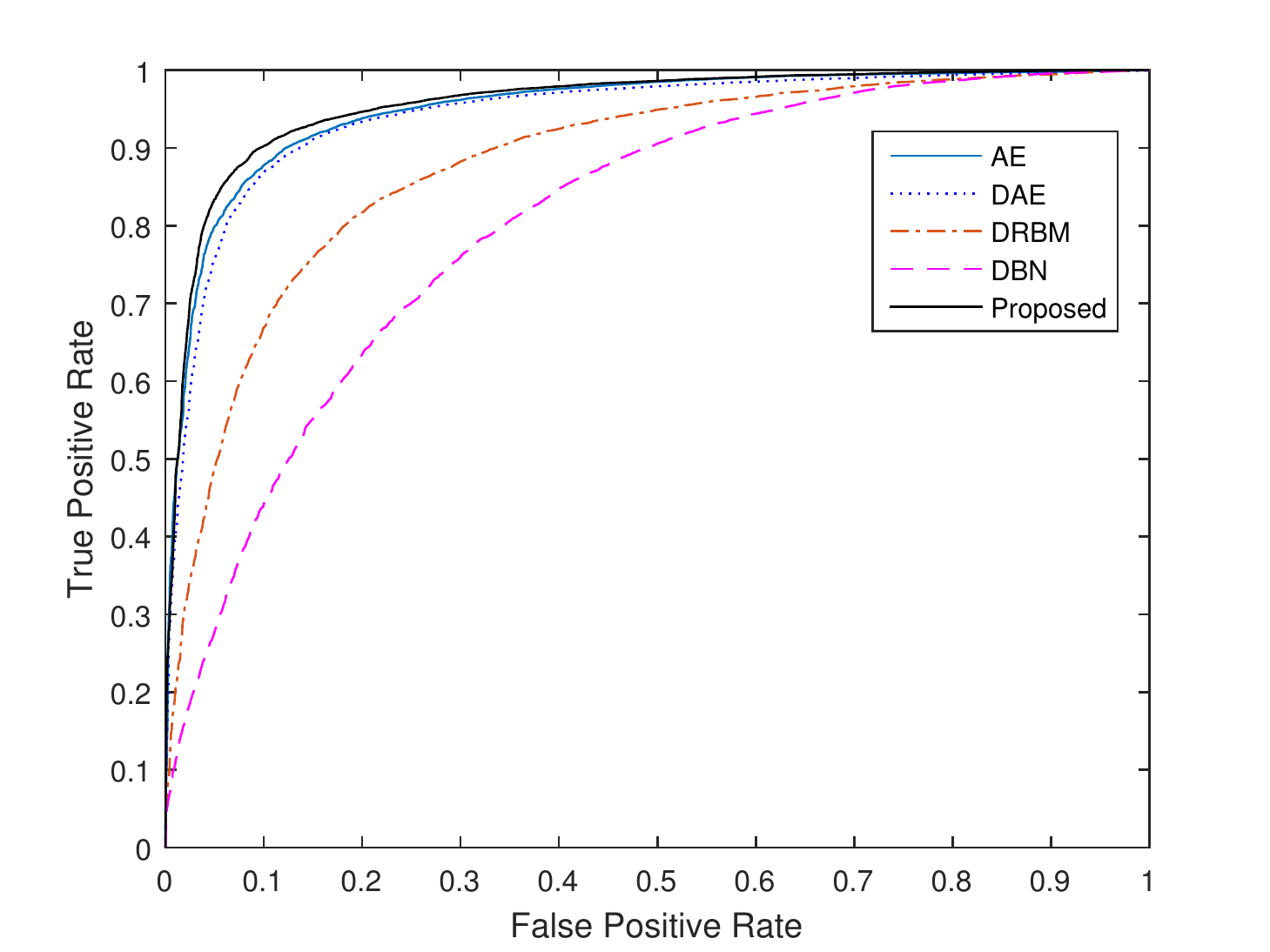}} 
\subfloat[PolyU-NIR]{\includegraphics[width= 3.6in]{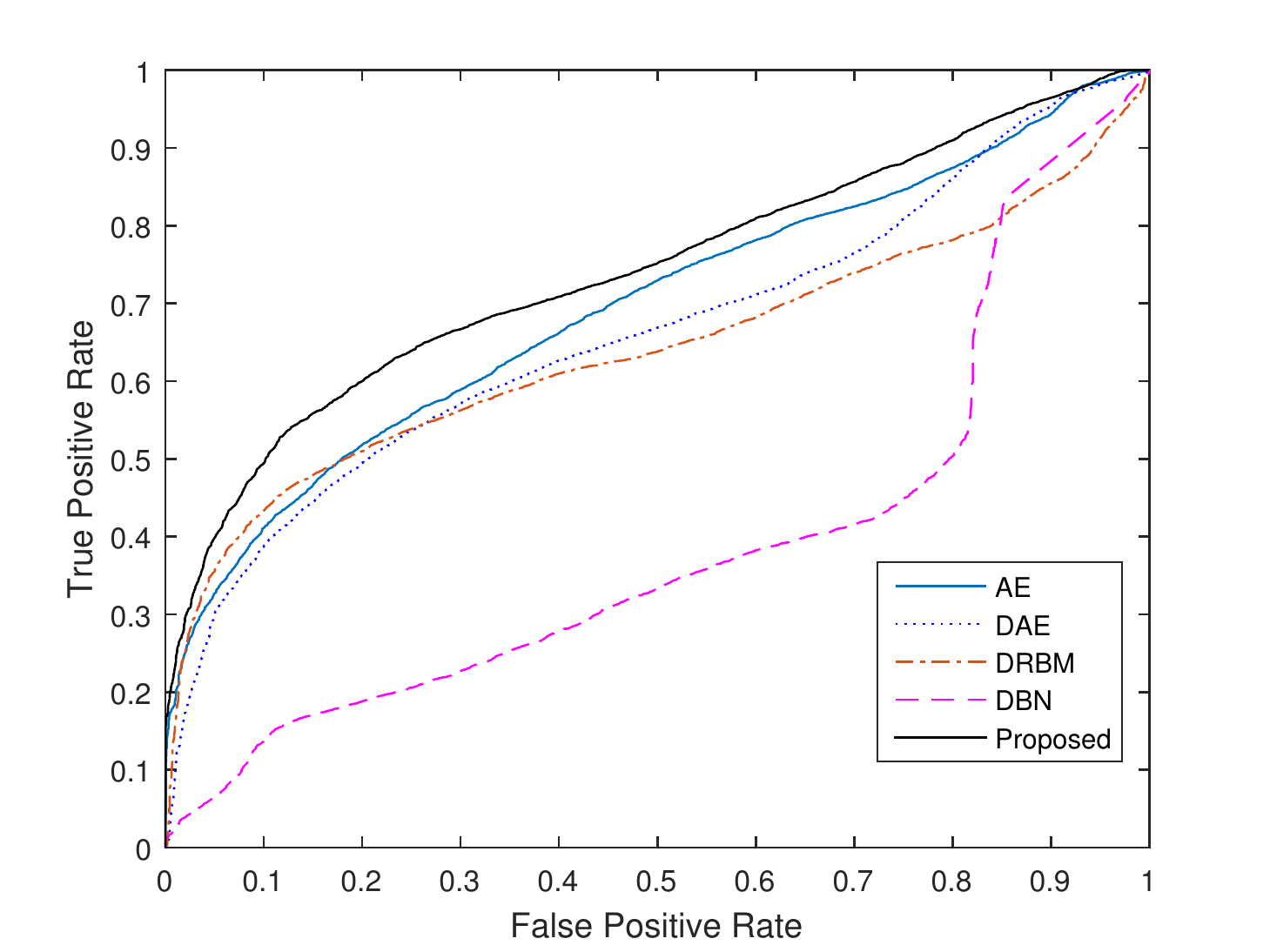}}  
\caption{Receiver Operating Characteristic (ROC) curves for $24 \times 24$ resolution face images. }
\label{fig:roc24}
\end{figure*}

\begin{figure*}[h]
\centering
\subfloat[CMU Multi-PIE]{\includegraphics[width= 3.6in]{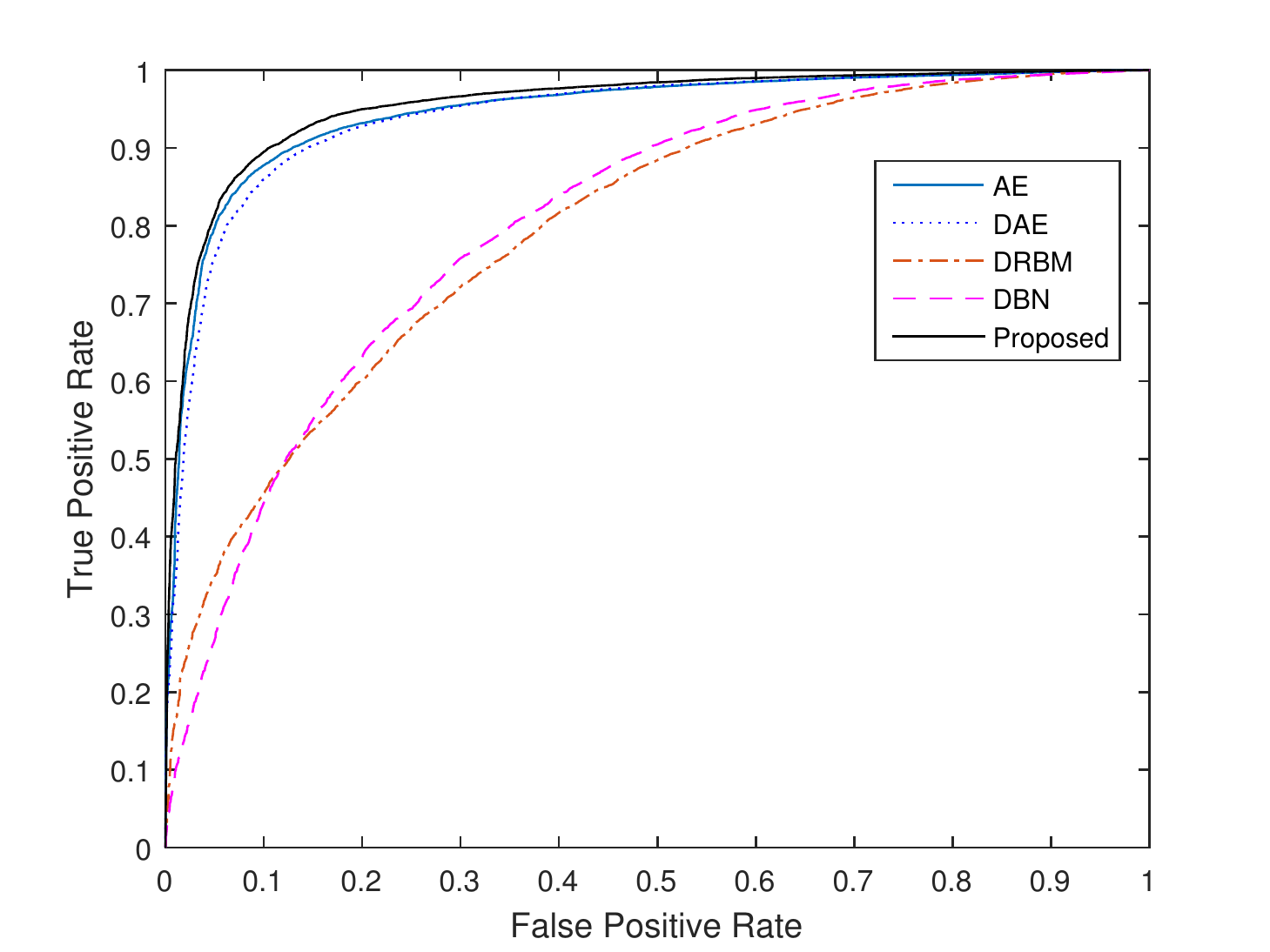}} 
\subfloat[PolyU-NIR]{\includegraphics[width= 3.6in]{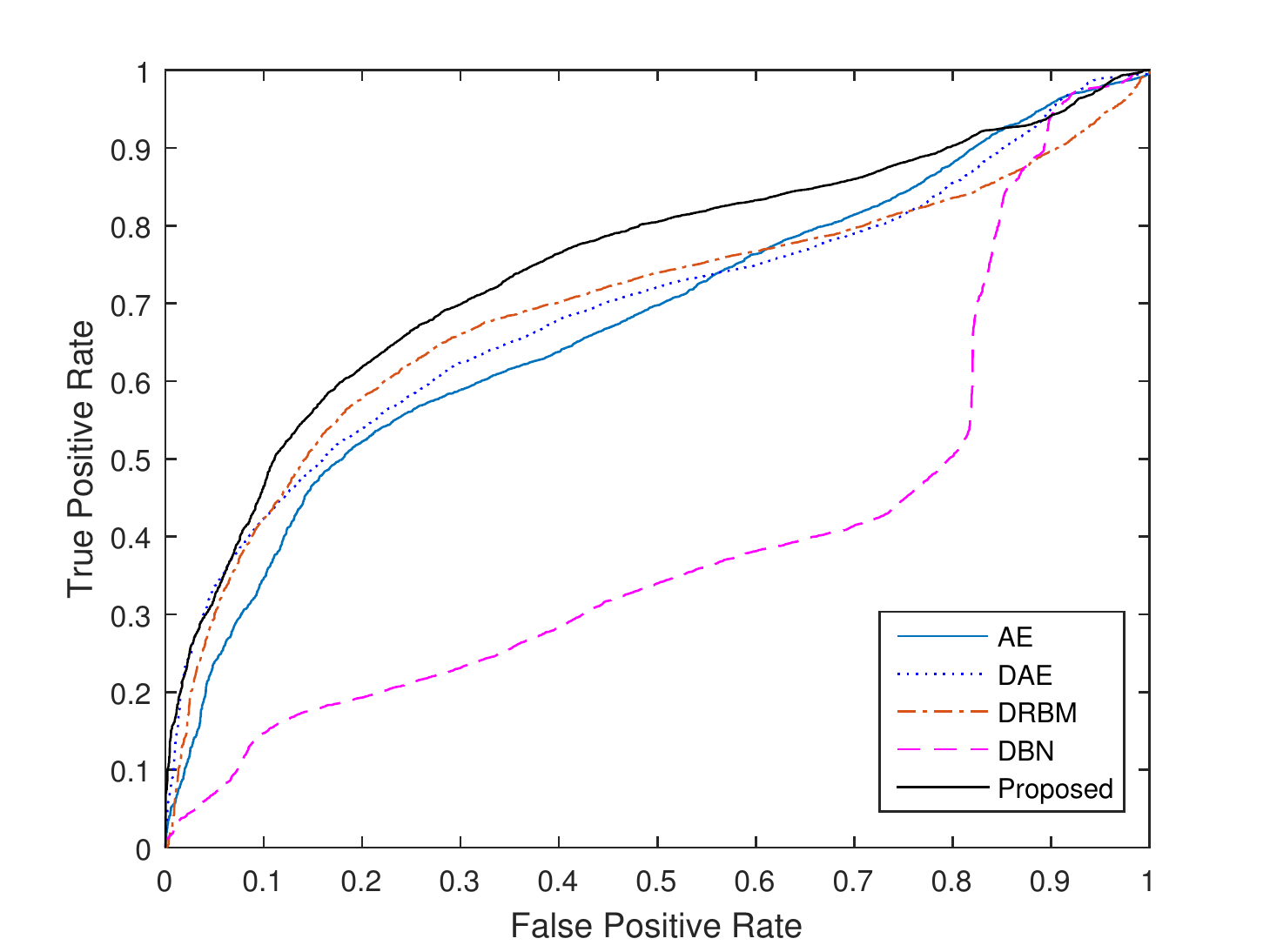}}  
\caption{Receiver Operating Characteristic (ROC) curves for $16 \times 16$ resolution face images. }
\label{fig:roc16}
\end{figure*}

$\mathbf{16\times16:}$ Table \ref{res16} presents the classification accuracies achieved by the models and commercial-off-the-shelf systems. For the visible spectrum (CMU Multi-PIE), a classification accuracy of 89.57\% is achieved, which depicts improvement over other comparative algorithms. It is important to note that the commercial matchers, Face++ and Luxand fail to process any image of this resolution, thereby resulting in 0\% classification performance. Experiments in the NIR spectrum showcase that the proposed model outperforms existing architectures by at least 5\%, resulting in a classification accuracy of 69.86\%. Fig. \ref{fig:roc24} and \ref{fig:roc16} present the Receiver Operative Characteristic (ROC) curves obtained for the experiments. 

\begin{figure}[h]
\centering
\includegraphics[trim={0cm 1.9cm 0cm 0cm},clip, width= 3.6in]{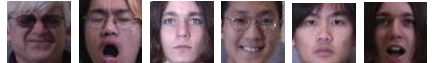} 
\caption{Sample male images misclassified as females by AutoGen. Most of the mis-classifications are categorized by unusual hairstyle, expression, or accessories such as sunglasses.}
\label{mis}
\end{figure}

Table \ref{classWise} also presents the class-wise classification accuracy obtained using AutoGen for the above experiments. It can be observed that for face images of resolution $24\times24$, female classification accuracy is significantly higher than male classification accuracy for both visible as well as NIR images. Upon analyzing the misclassified samples of males as females (Fig. \ref{mis}), it can be observed that unusual hairstyle, accessories, and sunglasses act as challenging artifacts for gender recognition. The performance of commercial-off-the-shelf systems, especially for $16\times16$ face images further reinstates the need for robust algorithms for gender recognition. It can also be observed that the accuracies obtained by AutoGen on $16\times16$ face images and $24\times24$ vary by less than 2\%, for each dataset. This suggests that even with low resolution face images, the model is able to learn sufficient discriminative features. 

\begin{table}
\renewcommand*{\arraystretch}{1.2}
\begin{center}
\centering
\caption{Classification accuracies (\%) for gender classification on $24\times24$ face images, for SCface dataset. }
\begin{tabular}{|M{8em}||M{7.5em}|M{7.5em}|} 
\hline
\textbf{Algorithm} & \textbf{Visible Spectrum} & \textbf{NIR Spectrum} \\
\hline
\hline
 AE & 84.29 & 91.83  \\ 
 \hline
DAE &  81.35 & 88.61  \\
\hline
DBN &  70.19 & 50.00 \\
\hline
Face++ &  2.72 & 0.00  \\
\hline
Luxand & 65.54 & 15.26 \\
\hline
\textbf{AutoGen} & \textbf{88.53} & \textbf{95.79} \\
\hline
\end{tabular}
\label{tab:scface}
\end{center}
\end{table}

\begin{figure}
\centering
\subfloat[CMU Multi-PIE]{\includegraphics[trim={0.5cm 0cm 0.5cm 0cm},clip, width= 3.5in]{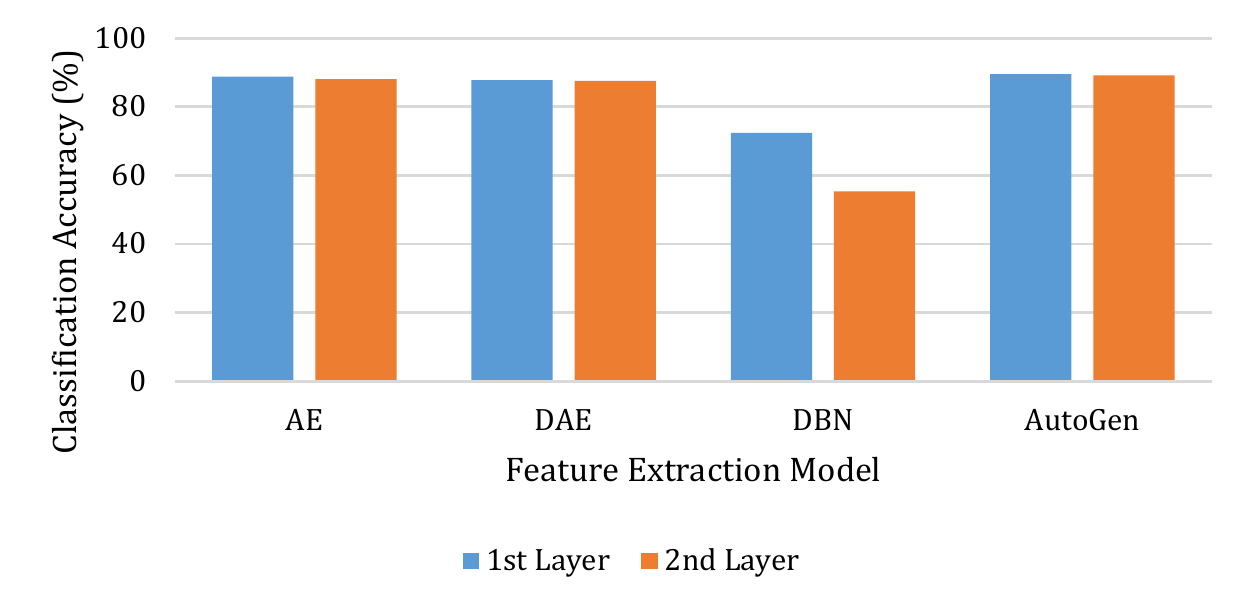}} \\ 
\subfloat[PolyU-NIR]{\includegraphics[trim={0.5cm 0cm 0.5cm 0cm},clip, width= 3.5in]{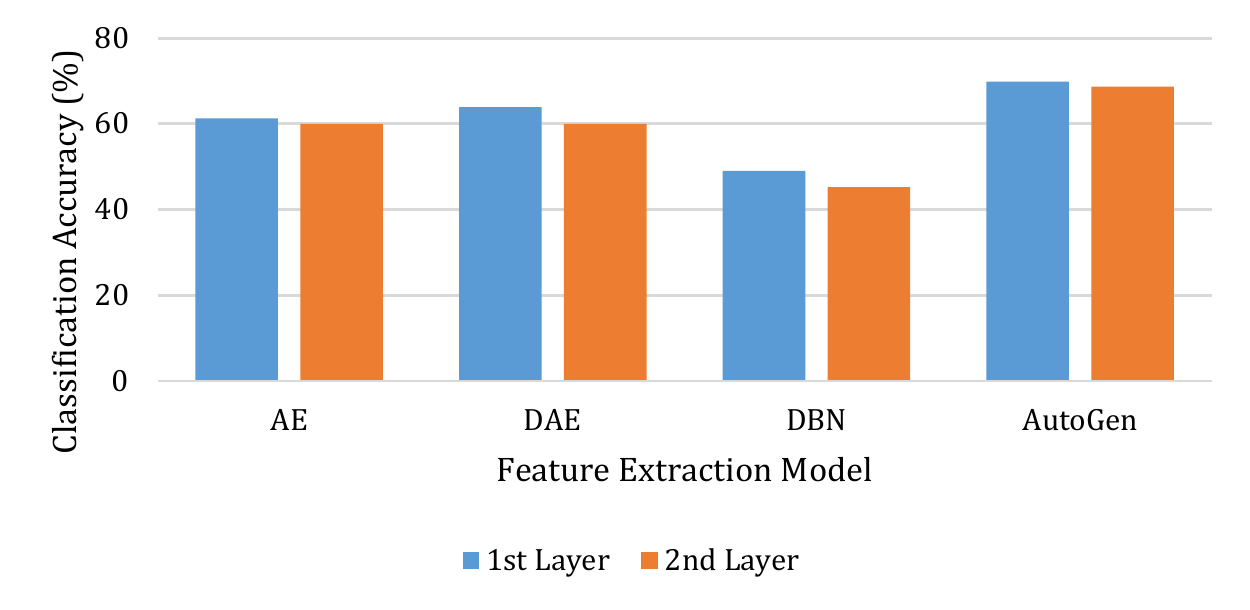}}  
\caption{Comparison of classification accuracies for two layer feature extraction models, for face images having $16\times16$ resolution. }
\label{deep}
\end{figure}

\subsection{Data Captured in in Real-World Scenarios: SCface Dataset}
SCface dataset \cite{scface} has been captured in real world conditions containing 130 subjects. Since the number of training samples for SCface dataset (both NIR and visible spectrum) are very less, a fine-tuning approach was applied for all models. The trained feature extractors for CMU Multi-PIE and PolyU-NIR were fine-tuned with the visible and NIR spectrum training set of SCface dataset, respectively. Table \ref{tab:scface} gives the classification accuracies obtained for both the spectra. It can be observed that AutoGen outperforms existing algorithms by at least 4\% for both the domains by obtaining a classification accuracy of 88.53\% and 95.79\%, respectively. Since the test samples for both the domains and feature extraction models vary in nature, direct comparison cannot be drawn between the accuracies of the two spectra. The improved performance of AutoGen with a small training set further motivates the usage of the proposed algorithm for the given task.  

\begin{figure}
\centering
\includegraphics[width=1\linewidth]{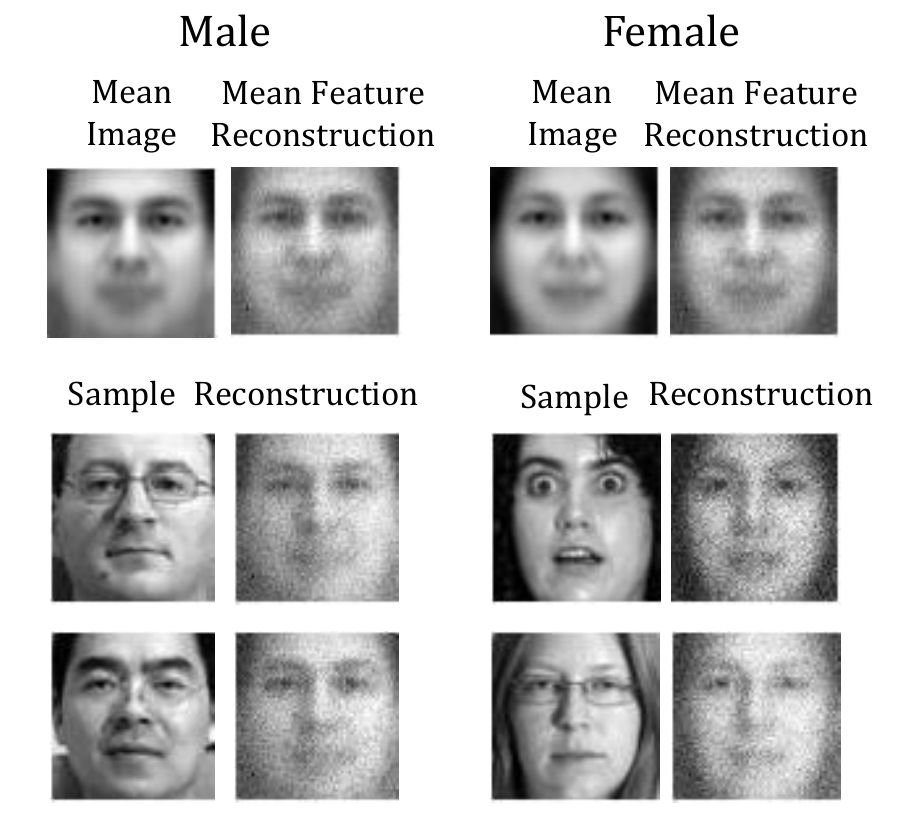}
\caption{Sample reconstructed images from CMU Multi-PIE dataset using AutoGen.}
\label{recon}
\end{figure}

\subsection{Training AutoGen for Spectrum Invariant Gender Classification}
In an attempt to model the gender variations across spectra in a single architecture, AutoGen was trained for face images of both spectrum, having a resolution of $16\times16$ pixels. A single model of AutoGen was trained using images of NIR and visible spectrum from the training sets of CMU Multi-PIE and PolyU-NIR. A classification accuracy of 88.47\% and 69.66\% is obtained for CMU Multi-PIE and PolyU-NIR face datasets respectively. In case of visible spectrum, a drop of around 1.1\% is observed in the recognition performance (88.47\% as opposed to 89.57\%), and a drop of less than a percent is observed in case of NIR gender classification. This experiment suggests that while a combined model does not display any improvement in the classification performances, however, the drop in accuracies is also very less. This motivates formulation of spectrum invariant models for the given task.

\subsection{Effect of Going Deeper}
To study the effect of deeper architectures for the purpose of gender classification, all feature extraction models were trained for two layers. Greedy layer-wise training algorithm \cite{greedy} was applied for $16\times16$ face images. Fig. \ref{deep} presents the classification accuracies obtained after learning two layers of feature extraction models, along with the first layer classification results. It can be observed that the performance of all models reduce upon going deeper, or upon learning higher level of abstractions. While the accuracy of the proposed model reduces by less than 1.5\%, the classification performance of DBN reduces by at most 17\% (for CMU Multi-PIE). This demonstrates that while AutoGen can be used for learning deeper feature representations, however, for the task of gender recognition in low resolution images, a single layer model performs best. 

Fig. \ref{recon} depicts some sample reconstructed images from CMU Multi-PIE dataset, obtained using AutoGen. It can be seen that the reconstructions of the mean male and female feature representations appear visually similar to the mean images. Along with that, the reconstructions of different samples demonstrate that the reconstructed face images are representative of the class to which the sample belongs. It can also be seen that the reconstructed image of a sample is closer to the mean reconstruction of the sample's class, as opposed to the other class. These visualizations along with the experimental results motivate the use of AutoGen for the task of gender classification on low resolution face images. 

\section{Conclusion}
This research proposes a novel Class Representative Autoencoder architecture for learning discriminative class-specific feature representations. The model, termed as \textit{AutoGen}, is presented for the task of multi-spectral low resolution gender recognition in face images. It aims to incorporate inter-class and intra-class variations at the time of feature learning, resulting in discriminative features. Experimental results on visible as well as NIR spectrum face images, along with the performance of spectrum invariant gender classification demonstrate the effectiveness of it's feature learning process. Comparison with existing algorithms and commercial-off-the-shelf systems for multiple datasets and different resolutions strengthens the need to incorporate class-specific discriminative information at the feature extraction stage, and motivates the use of AutoGen. 

\section{Acknowledgment}
This research is supported from a grant from Ministry of Electronics and Information Technology, India and Infosys Centre of Artificial Intelligence. S. Nagpal is supported through TCS PhD Fellowship. The authors acknowledge the feedback provided by the reviewers.


\bibliographystyle{IEEEtran}
\bibliography{bibFile}

\begin{thebibliography}{10}
\providecommand{\url}[1]{#1}
\csname url@samestyle\endcsname
\providecommand{\newblock}{\relax}
\providecommand{\bibinfo}[2]{#2}
\providecommand{\BIBentrySTDinterwordspacing}{\spaceskip=0pt\relax}
\providecommand{\BIBentryALTinterwordstretchfactor}{4}
\providecommand{\BIBentryALTinterwordspacing}{\spaceskip=\fontdimen2\font plus
\BIBentryALTinterwordstretchfactor\fontdimen3\font minus
  \fontdimen4\font\relax}
\providecommand{\BIBforeignlanguage}[2]{{%
\expandafter\ifx\csname l@#1\endcsname\relax
\typeout{** WARNING: IEEEtran.bst: No hyphenation pattern has been}%
\typeout{** loaded for the language `#1'. Using the pattern for}%
\typeout{** the default language instead.}%
\else
\language=\csname l@#1\endcsname
\fi
#2}}
\providecommand{\BIBdecl}{\relax}
\BIBdecl

\bibitem{gender1}
A.~Feingold, ``Gender differences in personality: a meta-analysis.''
  \emph{Psychological bulletin}, vol. 116, no.~3, p. 429, 1994.

\bibitem{gender2}
E.~E. Maccoby and C.~N. Jacklin, \emph{The psychology of sex
  differences}.\hskip 1em plus 0.5em minus 0.4em\relax Stanford University
  Press, 1974, vol.~1.

\bibitem{faceGender}
C.-B. Ng, Y.-H. Tay, and B.-M. Goi, ``A review of facial gender recognition,''
  \emph{Pattern Analysis and Applications}, vol.~18, no.~4, pp. 739--755, 2015.

\bibitem{irisGender}
J.~E. Tapia, C.~A. Perez, and K.~W. Bowyer, ``Gender classification from the
  same iris code used for recognition,'' \emph{IEEE Transactions on Information
  Forensics and Security}, vol.~11, no.~8, pp. 1760--1770, 2016.

\bibitem{fingerprintGender}
A.~Rattani, C.~Chen, and A.~Ross, ``Evaluation of texture descriptors for
  automated gender estimation from fingerprints,'' in \emph{European Conference
  on Computer Vision Workshop}, 2014, pp. 764--777.

\bibitem{keystrokeGender}
R.~Giot and C.~Rosenberger, ``A new soft biometric approach for keystroke
  dynamics based on gender recognition,'' \emph{International Journal of
  Information Technology and Management}, vol.~11, no. 1-2, pp. 35--49, 2012.

\bibitem{gaitGender}
J.~Lu, G.~Wang, and P.~Moulin, ``Human identity and gender recognition from
  gait sequences with arbitrary walking directions,'' \emph{IEEE Transactions
  on Information Forensics and Security}, vol.~9, no.~1, pp. 51--61, 2014.

\bibitem{scface}
M.~Grgic, K.~Delac, and S.~Grgic, ``S{C}face - {S}urveillance {C}ameras {F}ace
  {D}atabase,'' \emph{Multimedia Tools Application}, vol.~51, no.~3, pp.
  863--879, 2011.

\bibitem{moghaddam02}
B.~Moghaddam and M.-H. Yang, ``Learning gender with support faces,'' \emph{IEEE
  Transactions on Pattern Analysis and Machine Intelligence}, vol.~24, no.~5,
  pp. 707--711, 2002.

\bibitem{andreu14}
Y.~Andreu, J.~López-Centelles, R.~A. Mollineda, and P.~García-Sevilla,
  ``Analysis of the effect of image resolution on automatic face gender
  classification,'' in \emph{International Conference on Pattern Recognition},
  2014, pp. 273--278.

\bibitem{deepGender}
F.~Juefei-Xu, E.~Verma, P.~Goel, A.~Cherodian, and M.~Savvides, ``Deepgender:
  Occlusion and low resolution robust facial gender classification via
  progressively trained convolutional neural networks with attention,'' in
  \emph{Computer Vision and Pattern Recognition Workshops}, 2016.

\bibitem{ross11}
C.~Chen and A.~Ross, ``Evaluation of gender classification methods on thermal
  and near-infrared face images,'' in \emph{International Joint Conference on
  Biometrics}, 2011, pp. 1--8.

\bibitem{ross11_2}
A.~Ross and C.~Chen, ``Can gender be predicted from near-infrared face
  images?'' in \emph{International Conference on Image Analysis and
  Recognition}, 2011, pp. 120--129.

\bibitem{narang16}
N.~Narang and T.~Bourlai, ``Gender and ethnicity classification using deep
  learning in heterogeneous face recognition,'' in \emph{International
  Conference on Biometrics}, 2016, pp. 1--8.

\bibitem{sdae}
P.~Vincent, H.~Larochelle, I.~Lajoie, Y.~Bengio, and P.-A. Manzagol, ``Stacked
  denoising autoencoders: Learning useful representations in a deep network
  with a local denoising criterion,'' \emph{The Journal of Machine Learning
  Research}, vol.~11, pp. 3371--3408, 2010.

\bibitem{ng2011sparse}
A.~Ng, ``Sparse autoencoder,'' \emph{CS294A Lecture notes}, vol.~72, pp. 1--19,
  2011.

\bibitem{highContract}
S.~Rifai, G.~Mesnil, P.~Vincent, X.~Muller, Y.~Bengio, Y.~Dauphin, and
  X.~Glorot, ``Higher order contractive auto-encoder,'' in \emph{European
  Conference on Machine Learning and Knowledge Discovery in Databases}, 2011,
  pp. 645--660.

\bibitem{contrast}
X.~Zheng, Z.~Wu, H.~Meng, and L.~Cai, ``Contrastive auto-encoder for phoneme
  recognition,'' in \emph{International Conference on Acoustics, Speech and
  Signal Processing}, 2014, pp. 2529--2533.

\bibitem{super}
S.~Gao, Y.~Zhang, K.~Jia, J.~Lu, and Y.~Zhang, ``Single sample face recognition
  via learning deep supervised autoencoders,'' \emph{IEEE Transactions on
  Information Forensics and Security}, vol.~10, no.~10, pp. 2108--2118, 2015.

\bibitem{pamiMaj}
A.~Majumdar, R.~Singh, and M.~Vatsa, ``Face recognition via class sparsity
  based supervised encoding,'' \emph{IEEE Transactions on Pattern Analysis and
  Machine Intelligence}, 2016.

\bibitem{mpie}
R.~Gross, I.~Matthews, J.~Cohn, T.~Kanade, and S.~Baker, ``Multi-{PIE},''
  \emph{Image Vision Computing}, vol.~28, no.~5, pp. 807--813, 2010.

\bibitem{backProp}
R.~Hecht-Nielsen, ``Theory of the backpropagation neural network,'' in
  \emph{International Joint Conference on Neural Networks}, 1989, pp. 593--605.

\bibitem{greedy}
Y.~Bengio, P.~Lamblin, D.~Popovici, and H.~Larochelle, ``Greedy layer-wise
  training of deep networks,'' in \emph{Neural Information Processing
  Systems}.\hskip 1em plus 0.5em minus 0.4em\relax MIT Press, 2007.

\bibitem{polyu}
B.~Zhang, L.~Zhang, D.~Zhang, and L.~Shen, ``Directional binary code with
  application to {PolyU Near-infrared Face Database},'' \emph{Pattern
  Recognition Letters}, vol.~31, no.~14, pp. 2337--2344, 2010.

\bibitem{viola}
P.~Viola and M.~J. Jones, ``Robust real-time face detection,''
  \emph{International Journal Computer Vision}, vol.~57, no.~2, pp. 137--154,
  2004.

\bibitem{ae}
G.~Hinton and R.~Salakhutdinov, ``Reducing the dimensionality of data with
  neural networks,'' \emph{Science}, vol. 313, no. 5786, pp. 504 -- 507, 2006.

\bibitem{dae}
P.~Vincent, H.~Larochelle, Y.~Bengio, and P.-A. Manzagol, ``Extracting and
  composing robust features with denoising autoencoders,'' in
  \emph{International Conference on Machine Learning}, 2008, pp. 1096--1103.

\bibitem{dbn}
G.~E. Hinton, S.~Osindero, and Y.-W. Teh, ``A fast learning algorithm for deep
  belief nets,'' \emph{Neural Computing}, vol.~18, no.~7, pp. 1527--1554, 2006.

\bibitem{drbm}
H.~Larochelle and Y.~Bengio, ``Classification using discriminative restricted
  boltzmann machines,'' in \emph{International Conference on Machine Learning},
  2008, pp. 536--543.

\bibitem{luxand}
Luxand. https://www.luxand.com.

\bibitem{faceplusplus}
Face{++}. http://www.faceplusplus.com/.

\end{thebibliography}

\end{document}